\pdfoutput=1

\documentclass[11pt]{article}

\usepackage{ACL2023}

\usepackage{times}
\usepackage{latexsym}
\usepackage{graphicx}
\usepackage{bm,amsfonts,amssymb,amsmath}
\usepackage{tabularx} 
\usepackage{multicol}
\newcolumntype{C}{>{\centering\arraybackslash}X} 
\usepackage[T1]{fontenc}

\usepackage[utf8]{inputenc}

\usepackage{microtype}

\usepackage{inconsolata}

\usepackage[textsize=scriptsize]{todonotes}
\usepackage{xcolor}
\usepackage{booktabs}
\usepackage{multirow}
\usepackage{multicol}

\title{Language Models (Mostly) Do Not Consider Emotion Triggers \\ When Predicting Emotion}

\author{
    \textbf{Smriti Singh}$^1$\quad\textbf{Cornelia Caragea}$^2$\quad\textbf{Junyi Jessy Li}$^1$\\
    $^1$The University of Texas at Austin\\$^2$University of Illinois Chicago\\
    \texttt{\{smritisingh26, jessy\}@utexas.edu, cornelia@uic.edu}
}

\begin{document}
\maketitle
\begin{abstract}
Situations and events evoke emotions in humans, but to what extent do they inform the prediction of emotion detection models?  This work investigates how well human-annotated emotion triggers correlate with features that models deemed salient in their prediction of emotions. First, we introduce a novel dataset {\sc EmoTrigger}, consisting of 900 social media posts sourced from three different datasets; these were annotated by experts for emotion triggers with high agreement. Using {\sc EmoTrigger}, we evaluate the ability of  large language models (LLMs) to identify emotion triggers, and conduct a comparative analysis of the features considered important for these tasks between LLMs and  fine-tuned models. Our analysis reveals that emotion triggers are largely not considered salient features for emotion prediction models, instead there is intricate interplay between various features and the task of emotion detection.
\end{abstract}

\section{Introduction}\label{sec:intro}

Understanding perceived emotions and how they are expressed can be immensely useful for providing emotional support, sharing joyful situations, or in a therapy session, thus emotion detection has become a well-studied task~\cite{strapparava-mihalcea-2007-semeval,Wang2012HarnessingT,abdul-mageed-ungar-2017-emonet,khanpour-caragea-2018-fine,liu-etal-2019-dens,sosea-caragea-2020-canceremo,demszky-etal-2020-goemotions,desai-etal-2020-detecting,sosea-caragea-2021-emlm,sosea-etal-2022-emotion,hosseini-caragea-2022-calibrating,Hosseini_Caragea_2023,hosseini-caragea-2023-semi}. However, though existing work has sought to identify what triggers or causes a particular emotion~\cite{lee-etal-2010-emotion,chen-etal-2010-emotion,gui-etal-2016-event,xia-ding-2019-emotion,zhan-etal-2022-feel,sosea-etal-2023-unsupervised}, the relationship between those triggers and the prediction of emotion detection models is little understood. This relationship is crucial to investigate, without which the interpretation of perceived emotions---or claims for a model to be able to do so---is \emph{hollow}~\cite{james1948emotion}.

While humans can intuitively construe emotional reactions with events that trigger them~\cite{jie2023meta}, it is unclear to what extent, if any, current NLP models are doing so.
Prior work trained models to \emph{learn} to recognize and summarize emotion triggers or causes. In this work, we instead ask the question: \emph{what roles do emotion triggers play in emotion prediction?} In addition to fine-tuned transformer models shown to be performant on this task, we additionally put an emphasis on large language models (LLMs) including both API-based and open-sourced ones, since their capability for trigger prediction has not been explored. 

To ground our analysis, we present {\sc EmoTrigger}, a linguist-annotated dataset of emotion triggers (as extractive text spans), over three social media corpora with labeled emotions across different themes: CancerEmo~\cite{sosea-caragea-2020-canceremo}, HurricaneEmo~\cite{desai-etal-2020-detecting}, and GoEmotions~\cite{demszky-etal-2020-goemotions}. This is, to the best of our knowledge, the first dataset annotated with high-quality triggers focusing on \emph{short} social media texts. 
Engaging with tools that attribute model prediction~\cite{lundberg2017unified} and prompts that elicit natural language explanations from LLMs, we
aim to answer the following research questions:

\vspace{-0.5em}
\begin{enumerate}
\itemsep-0.3em 
    \item Are LLMs capable of detecting emotions and  identifying their triggers? 
    \item To what extent do emotion prediction models rely on features that reflect emotion triggers?
    \item How often do the triggers overlap with keyphrases or emotion words? \vspace{-0.5em}
\end{enumerate}

We find that LLMs can identify emotions with high accuracy, but the performance for identifying triggers is mixed. With the exception of GPT-4, word features deemed salient for emotion prediction are only marginally related to these triggers. Instead, we found that automatically extracted keyphrases~\cite{bougouin-etal-2013-topicrank} are highly correlated with salient features. 
Overall, we establish that the current state of the art language models cannot proficiently construe emotional reactions with events that trigger them.

We release {\sc EmoTrigger} at 
\url{https://github.com/smritisingh26/EmoTrigger}.

\section{The {\sc EmoTrigger} Dataset}
We first present {\sc EmoTrigger}, a dataset annotated with ground-truth emotion triggers. {\sc EmoTrigger} consists of subsets from three well-known datasets commonly used in emotion prediction, on top of which we engage expert annotators to highlight triggers for the annotated emotion.
Although prior emotion cause datasets exist, they largely focus on explicit emotions~\cite{chen-etal-2010-emotion,gui-etal-2016-event}. \citet{sosea-etal-2023-unsupervised}'s dataset for Reddit posts involves multi-sentence triggers for summarization, while this work focuses on word or phrase level attributions and explanations. 

\paragraph{Source Datasets}
(1) \textit{HurricaneEmo}~\cite{desai-etal-2020-detecting} consists of tweets related to 3 hurricanes.
Each tweet is annotated for 24 fine-grained emotions in Plutchik's wheel of emotions~\cite{doi:10.1177/053901882021004003}.
We map the emotions onto Plutchik's 8 basic emotions (Appendix~\ref{sec:emomapping}).

(2) \emph{CancerEmo}~\cite{sosea-caragea-2020-canceremo} consists of 
sentences sampled from an online cancer survivors network. This dataset is annotated with Plutchik's 8 basic emotions.

(3) \emph{GoEmotions}~\cite{demszky-etal-2020-goemotions} is a general domain dataset that consists of 
sentences extracted from popular English subreddits, labeled for 27 emotion categories or Neutral, which are mapped to the coarser Ekman's 6 emotions~\cite{ekman1992argument}.

To balance the emotions in {\sc EmoTrigger}, we sample 20 examples per emotion from each constituent dataset. In total, we have 900 samples in our dataset, 160 each from HurricaneEmo and CancerEmo, and 580 samples from GoEmotions. We make sure our samples do not contain links or images, i.e., the text stands alone from other media.

\vspace{-0.5em}
\paragraph{Annotation}
Our annotation team consists of three linguistics undergraduates experienced with emotion-related text annotation tasks, who annotate the 900 instances (tweets or sentences) for emotion triggers. The annotation task is to find the emotion triggers for a given text and the gold label emotion. The annotators were paid at \$15/hr. 
Our annotation instructions are listed in Appendix~\ref{ann_instr}. 

\begin{figure}[t]
  \centering
  \includegraphics[width=0.8\linewidth]{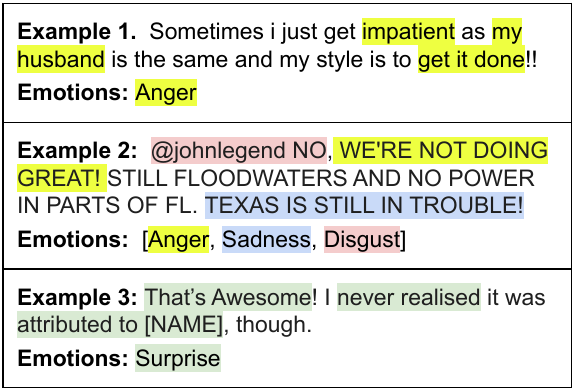} 
  \vspace{-0.5em}
  \caption{Examples from {\sc EmoTrigger}. Triggers highlighted with the same color as the respective emotions.}
  \label{fig:intro}
\end{figure}

We report token-level inter-annotator agreement with Fleiss Kappa~\cite{fleiss1971measuring}, yielding an average of 0.91, indicating substantial to perfect agreement~\cite{artstein-poesio-2008-survey}. Table~\ref{tab:ann} shows the agreement between the annotators, per dataset. This is a positive indication that humans can reliably identify triggers of an emotion in short texts. Examples of {\sc EmoTrigger} are shown in Figure~\ref{fig:intro}.

\begin{table}
\centering
\small
\begin{tabular}{c|ccc|c}
\toprule
\textbf{Subset}      & \textbf{A1/A2} & \textbf{A1/A3} & \textbf{A2/A3} & \textbf{Avg} \\ \midrule
\textbf{HurricaneEmo} &  0.92           & 0.89           & 0.92           & 0.91            \\
\textbf{CancerEmo}    & $0.93$          & $0.89$           & $0.91$           & $0.92$           \\
\textbf{GoEmotions}   & $0.91$          & $0.88$           & $0.93$           & $0.90$           \\ \bottomrule
\end{tabular}
\caption{Annotator agreement (token-level Fleiss Kappa) across subsets of {\sc EmoTrigger}.}
\vspace{-3mm}
\label{tab:ann}
\end{table}

\section{Study Design}

We study to what extent words most informative to the final model predictions constitute triggers (as annotated in {\sc EmoTrigger}) of the emotion they are predicting. This section describes our strategy to do so for each model type.

\vspace{-0.5em}
\paragraph{LLMs}
We experimented with 
GPT-4, Llama2-Chat-13B~\cite{touvron2023llama}, and Alpaca-13B~\cite{alpaca}. Our main LLM prompt asks the model to predict emotions present in a given piece of text, and elicit explanations as words most important for detecting those emotions; annotated triggers were used as ``salient features''. Due to the poor zero-shot performance (Appendix Table~\ref{tab:task1-0shot}), we report few-shot results (prompt in Appendix \ref{prompt1}).\footnote{
We found that the prompt used for GPT4 and LLama2Chat does not allow Alpaca to engage with the text properly, thus we have a different prompt for Alpaca, as shown in \ref{prompt-alpaca}. Results from the old vs.\ new prompt are illustrated in Figure \ref{fig:alpaca-prompts} in the Appendix. To account for the spelling mistakes that some LLMS make, we compute the Levenshtein distance of words that have more than 4 letters and consider words that have a distance of less than or equal to 2.} 

Since there has been no formal investigation yet for LLM's ability to predict emotion triggers, we also include an \emph{oracle} that asks the model to identify triggers given the gold emotion label in a few-shot manner (prompt in Appendix~\ref{prompt2}). Our experiments are run using A100s available on Google Colab and take a total of approximately 50 hours.

\begin{figure}[t]
  \centering
  \includegraphics[width=1\linewidth]{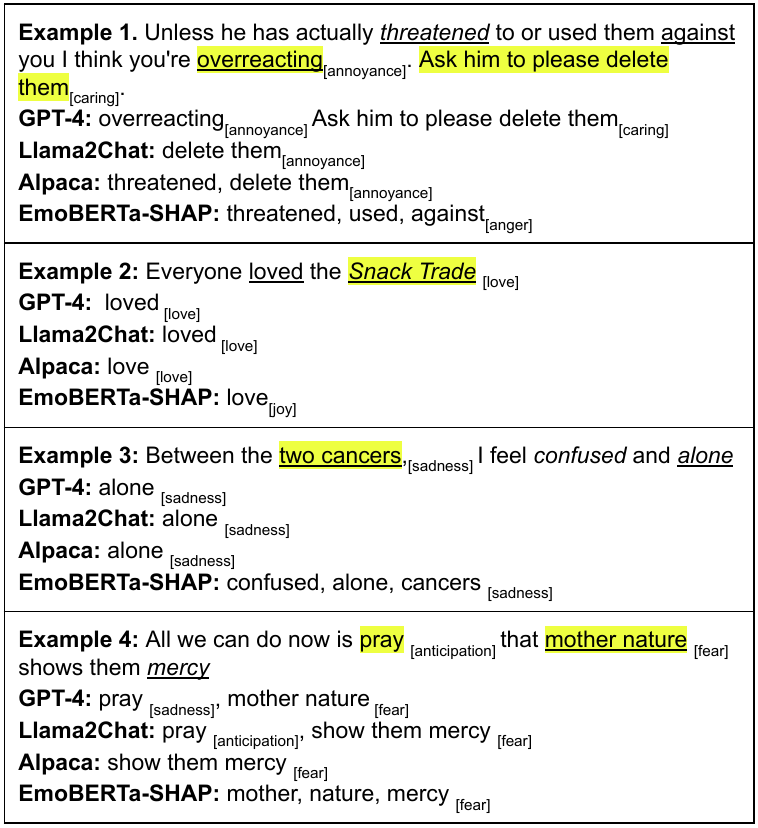} 
  \vspace{-1.5em}
  \caption{Examples of trigger identification. Gold label triggers are highlighted, keyphrases are underlined and EmoLex words are in italics. The emotions are in subscript to the triggers that caused them.}
 \vspace{-3mm}
  \label{fig:trigs}
\end{figure}

\vspace{-0.5em}
\paragraph{Fine-tuned Transformers}
We fine-tuned transformer models on the aforementioned datasets for emotion prediction, with EmoBERTA~\cite{kim2021emoberta} as the most performant model which we will use in subsequent analyses.\footnote{Other models include BERT~\cite{devlin2018bert}, DistilBERT~\cite{liu2019roberta}, RoBERTa~\cite{sanh2019distilbert}, DeBERTa~\cite{he2021deberta}. Their performance is summarized in Appendix Table~\ref{tab:transfomers}} To train these models, we use a 75-15-10 split for each dataset respectively. Hyperparameters listed in Appendix~\ref{app:hyperparams}.

To obtain salient features used for emotion prediction, we used SHAP~\cite{lundberg2017unified}, a model-agnostic method based on Shapley values that assigns an importance score to each feature (i.e., word) in the model prediction. Features with positive SHAP values positively impact the prediction, while those with negative values have a negative impact. The magnitude is a measure of how strong the effect is.

\vspace{-0.5em}
\paragraph{Unsupervised Comparators} We further consider two methods that extract key information in an unsupervised manner.
First, we use \textbf{EmoLex}~\cite{Mohammad13} to check to what extent human-annotated triggers or the salient features extracted by models correspond to words that express emotion themselves.
Second, we use \textbf{keyphrases} extracted from TopicRank~\cite{bougouin-etal-2013-topicrank}, to gauge to what extent corpora-specific keyphrases and themes correspond to triggers or salient features. TopicRank is an algorithm that performs graph-based keyphrase extraction by leveraging the topical representation of the document. The algorithm clusters keyphrases into topics and then utilizes a graph-based ranking model to assign a significance score to each topic, of which the candidates from top ranked topics become the extracted keyphrases. 

\begin{table}[t]
\centering
\small
\begin{tabular}{ccccc}
\toprule
& & & & \textbf{Emo-}\\
\textbf{Dataset}      & \textbf{GPT4} & \textbf{Llama2} & \textbf{Alpaca} & \textbf{BERTa} \\ \midrule
\textbf{HurricaneE.} & 0.920         & 0.851  & 0.756 & 0.483              \\ 
\textbf{CancerEmo}    & 0.914         & 0.842  & 0.723 & 0.378            \\ 
\textbf{GoEmotion}   & 0.907         & 0.820  & 0.707 & 0.341          \\ \bottomrule
\end{tabular}
\vspace{-0.5em}
\caption{ Macro F1 score for emotion detection. All scores are based on results of few-shot prompting.}
\label{tab:task2a}
\vspace{-1em}
\end{table}

\begin{table*}[t]
\centering
\small
\begin{tabular}{c|ccc|ccc|ccc}
\toprule
& \multicolumn{3}{c|}{\textbf{HurricaneEmo}} & \multicolumn{3}{c|}{\textbf{CancerEmo}} & \multicolumn{3}{c}{\textbf{GoEmotions}} \\
\textbf{Model} & F1 & ExactM & PartialM & F1 & ExactM & PartialM & F1 & ExactM & PartialM \\
\midrule
\textbf{GPT4 (known emotion)} & 0.7 & 0.38 & 0.9 & 0.71 & 0.39 & 0.91 & 0.68 & 0.33 & 0.9 \\
\textbf{Llama2 (known emotion)} & 0.31 & 0.11 & 0.35 & 0.3 & 0.11 & 0.37 & 0.28 & 0.09 & 0.34 \\
\textbf{Alpaca (known emotion)} & 0.23 & 0.12 & 0.29 & 0.2 & 0.09 & 0.25 & 0.19 & 0.06 & 0.21 \\ \midrule
\textbf{GPT4} & 0.66 & 0.35 & 0.87 & 0.68 & 0.37 & 0.88 & 0.65 & 0.31 & 0.89 \\
\textbf{Llama2} & 0.27 & 0.09 & 0.27 & 0.28 & 0.08 & 0.29 & 0.25 & 0.08 & 0.31 \\
\textbf{Alpaca} & 0.24 & 0.08 & 0.25 & 0.25 & 0.08 & 0.28 & 0.23 & 0.06 & 0.28 \\
\textbf{EmoBERTA-SHAP} & 0.21 & 0.08 & 0.23 & 0.19 & 0.09 & 0.18 & 0.18 & 0.07 & 0.19 \\ \midrule
\textbf{Keyphrases} & 0.19 & 0.08 & 0.23 & 0.2 & 0.08 & 0.25 & 0.18 & 0.08 & 0.18 \\
\textbf{Emolex} & 0.08 & 0 & 0.07 & 0.05 & 0 & 0.05 & 0.06 & 0.01 & 0.07 \\
\bottomrule
\end{tabular}
\vspace{-0.5em}
\caption{Macro-F1, exact and partial match  to assess the overlap between salient words and annotated triggers.}
\label{tab:task1}
\end{table*}

\begin{table*}[h]
\centering
\small
\begin{tabular}{c|cc|cc|cc|cc}
\toprule
& \multicolumn{2}{c|}{\textbf{GPT-4 \&}} & \multicolumn{2}{c|}{\textbf{Llama2 \&}} & \multicolumn{2}{c|}{\textbf{Alpaca \&}} & \multicolumn{2}{c}{\textbf{EmoBerta-SHAP \&}} \\
\textbf{Dataset} & \textbf{Keyphrase} & \textbf{EmoLex} & \textbf{Keyphrase} & \textbf{EmoLex} & \textbf{Keyphrase} & \textbf{EmoLex} & \textbf{Keyphrase} & \textbf{EmoLex} \\ \midrule
\textbf{HurricaneEmo} & 0.948  & 0.401 & 0.655  & 0.311 & 0.633 & 0.297 & 0.963 & 0.375 \\
\textbf{CancerEmo} & 0.863  & 0.711 & 0.703  & 0.366 & 0.686 & 0.344 & 0.823 & 0.635 \\
\textbf{GoEmotions} & 0.875  & 0.717 & 0.720  & 0.312 & 0.711 & 0.300 & 0.855 & 0.707 \\ \bottomrule
\end{tabular}
\vspace{-0.5em}
\caption{Pearson's correlation values between words each model deems salient and extracted keyphrases or EmoLex words. The scores shown here are the computed average of scores across individual emotion classes.}
\label{tab:RQ3}
\end{table*}

\section{Results}
For binary token-level comparison of annotated triggers vs.\ words the models identified as important for emotion detection (henceforth ``\emph{salient features}''), we look at: \textbf{(1)} Word-level macro-averaged F1 comparing salient features against the union of annotator-highlighted triggers. 
\textbf{(2)} Exact match and partial match. We define an exact match when the model output matches at least one of the annotated triggers. We consider a subset of any annotated triggers as a partial match. 
With respect to EmoBERTa, we took a cutoff for the SHAP values: 0.85 for an exact match and 0.60 for a partial match. These values were obtained empirically on a randomly sampled set of size 200.
This set is sampled outside {\sc EmoTrigger}.

We also report the Pearson's correlation coefficients between SHAP values, Emolex words 
and the weights of generated keyphrases. 
For EmoLex and salient features from LLMs, we use a binary vector representation.

\paragraph{Are LLMs capable of detecting emotions and  identifying their triggers?}
We first show each models' performance on the emotion prediction task across all datasets in Table~\ref{tab:task2a}. In Appendix Table~\ref{tab:EmotionWise} we show per-emotion results. LLM performance is \emph{consistent} across emotions and datasets while EmoBERTa's performance is substantially inferior. Among the LLMs, GPT-4 outperforms open-sourced ones. 

The first portion of Table~\ref{tab:task1} reports LLM performance when specifically prompted to identify triggers given the annotated emotions. Per-emotion results can be found in Tables \ref{tab:Trigs-GPT4}, \ref{tab:Trigs-Llama2}, and \ref{tab:Trigs-Alpaca} in the Appendix. Again, GPT-4 is the most performant; Llama2 slightly outperforms Alpaca.

As seen in Table~\ref{tab:EmotionWise}, we find that GPT4 predicts emotions `fear' and `joy' with the highest F1-scores consistently across datasets, and struggles with `anticipation'. 
With LLama2 and Alpaca, per-emotion performance is largely dataset dependent.

While there is no emotion for which any model can consistently identify the triggers most accurately, it is worth noting that for GPT-4 and Llama2Chat, the lowest scores are corresponding to the identification of triggers for the emotion `anticipation'. This is reflected in Tables \ref{tab:Trigs-GPT4}, \ref{tab:Trigs-Llama2}, and \ref{tab:Trigs-Alpaca} in the Appendix. 

\paragraph{To what extent do emotion prediction models rely on features that reflect emotion triggers?}
The bottom portion of Table~\ref{tab:task1} shows how much the salient features for each model overlap with annotated emotion triggers. Salient LLM features align less well with triggers than the ``oracle'' scenario above, but the differences are within 3\%. Note that this is with few-shot prompting, since we observed much lower alignment with zero-shot (Appendix Table~\ref{tab:task1-0shot}).
With the exception of GPT-4, salient features in neither Llama2, Alpaca nor EmoBERTa-SHAP align with annotated triggers with very little exact match and partial match.

\paragraph{How often do the triggers overlap with keyphrases or emotion words?}
We further hypothesize that keyphrases in the dataset, as well as explicit emotion words, might align with salient features that models pick up. Table~\ref{tab:RQ3} tabulates the average Pearson’s correlation coefficients between salient features (in the case of SHAP, feature salience values) and keyphrase weights or EmoLex.

Surprisingly, the correlations with keyphrases are much higher than with EmoLex for all models and across datasets. This indicates that models rely on explicit emotion words to a lesser extent than expected, and keyphrases help characterize this discrepancy. This is especially true for fine-tuned model (EmoBERTa) on themed datasets (HurricaneEmo, CancerEmo): the SHAP values are much more correlated with EmoLex especially in GoEmotions. We find that emotions like Anger, Joy and Sadness are expressed more through EmoLex words, whereas emotions like Anticipation, Fear and Disgust are expressed more through keyphrases. This is reflected in Figure \ref{fig:features}. 

We observe that GPT4, LLama2 and Alpaca rely on keyphrases an average of 27.3\%, 35.9\%  and 36.3\% more than Emolex words respectively, whereas EmoBERTa relies on keyphrases an average of 40.8\% more than Emolex words. This is also reflected in Table \ref{tab:RQ3}. 

\section{Qualitative Analysis}

In this section, we provide details about 
what we observe in our analysis, in terms of 
what large language models get right, and what they get wrong. We also delve deeper into the comparison of the results of the experiments with LLMs and fine-tuned models.

\subsection{Trigger identification}\label{app:analysis:triggers}

When it comes to trigger identification, we find that GPT4 is generally good at identifying the right triggers, except for when it comes to comments about specific experiences or entities. In these cases, it confuses emotion words for emotion triggers. An example of this is provided in Figure~\ref{fig:trigs}. 

The distinction between the performance of GPT4 and the other models we evaluate is quite clear. We find that Llama2-chat struggles with identifying triggers even for simple sentences. Further, we observe that Llama2 makes spelling mistakes when it does identify the right triggers. Alpaca's performance is slightly worse than Llama2. Examples are demonstrated in Figure~\ref{fig:trigs}.

\subsection{LLM's ``attribution'' of its own predictions}\label{app:analysis:attr}

Here, we observe that the LLMs detect emotions with a significantly higher accuracy than transformer models. As shown in Figure \ref{fig:emo-detect}, we find that LLMs struggle with identifying all emotions correctly if multiple emotions are present.

However, a drop in accuracy can be observed in trigger identification when the gold label emotions are not provided. We observe that GPT4 identifies triggers for emotions that it detects, even when the emotions themselves are incorrect. Further, we find that even when it does get the emotions right, it sometimes chooses triggers differently (and sometimes incorrectly) compared to the ones it chose when the same emotions were provided to it in the prompt. We find that Llama2 and Alpaca exhibit similar behavior. Examples are given in Figure \ref{fig:features}.

\subsection{Understanding feature importance: contrasting LLMs with transformer models}
\label{app:analysis:understand}
We find that large language models are able to detect emotions significantly more accurately than their traditionally fine-tuned counterparts. As demonstrated in Figure \ref{fig:features}, we see that the LLMs that make correct prediction of emotion correctly identify keyphrases and EmoLex words (when present). We observe that this is not true with respect to EmoBERTa. Even though there is a very high correlation of SHAP values and keyphrases, it doesn't entail that EmoBERTa is able to detect the correct emotion. This indicates that paying attention to a single feature is not enough, and that a fundamental understanding of grammar and language may be necessary to perform emotion detection correctly. 

\section{Discussion and Conclusion}
In this paper, we present {\sc EmoTrigger}, a linguist-annotated dataset of emotion triggers (as extractive text spans), over three social media corpora with labeled emotions across different themes. We use this dataset to analyze what role emotion triggers play in emotion detection. 

Overall, we believe this work provides evidence that with the exception of very large models like GPT-4 (few-shot), open-sourced ones like Llama2-chat and Alpaca do not have a good understanding of what triggers an emotion. The finding that salient features correlate substantially with keyphrases, rather than emotion triggers, means that models are better at picking up corpus-level topical cues rather than possessing a deep understanding of emotions \emph{per se} as humans do. In Psychology, emotion is viewed as triggered by subjective evaluations (or appraisals) of particular events~\cite{zhan-etal-2022-feel,moors2013appraisal}; thus future work on more sophisticated emotional support open-source language models should address this flaw.

\section{Limitations}
In our work, we analyze what role emotion triggers play in emotion detection. While we believe the development and analysis of the {\sc EmoTrigger} dataset is a step forward in this area of research, our study has a few limitations. First, our dataset is relatively small in size, owing to the labor intensive process of human annotation and the consideration of computational expenses of using the data with LLMs. Second, we run our study on a limited number of LLMs. This is also due to the consideration of computational resources. Finally, our study only deals with text that is in English; we leave multilingual pursuits for future work.

\section*{Acknowledgements}
This research is partially supported by National Science Foundation (NSF) grants IIS-2107524 and IIS-2107487. We thank Keziah Kaylyn Reina, Kathryn Kazanas and Karim Villaescusa F. for their work on the annotation of {\sc EmoTrigger}. We also thank our reviewers for their insightful comments.

\bibliography{anthology,custom}
\bibliographystyle{acl_natbib}

\clearpage
\appendix

\section{Mapping of HurricaneEmo emotions}
\label{sec:emomapping}

We aggregate the classes present in the dataset such that the final dataset consists of tweets annotated for Plutchik's eight primary emotions— anger, fear, sadness, disgust, surprise, anticipation, trust, and joy. We also include a none class. This aggregation is shown below:
\begin{itemize}
    \item Anger:  anger, annoyance, rage
    \item Fear:  fear, apprehension, terror
    \item Sadness: sadness, grief, pensiveness
    \item Disgust: disgust, loathing, boredom
    \item Surprise: surprise, amazement, distraction
    \item Anticipation: anticipation, interest, vigilance 
    \item Trust: trust, admiration, acceptance
    \item Joy: joy, serenity, ecstasy
\end{itemize}

\section{Annotation Instructions}
\label{ann_instr}

Hello, and welcome to this annotation task! This is a task designed to understand the importance of different words in identifying emotions and their triggers. To complete it, we will be using Google Sheets. As a part of this task, you will be given a google sheet consisting of a sentence, each already annotated for several different emotions. You will complete the task of identifying emotion triggers, as stated below:

\paragraph{Identifying Emotion Triggers}
For this task, your assignment is to analyze the sentence and the annotated emotions, and identify the words/phrases that indicate the triggers of the emotions reflected by the gold label annotation. Triggers tend to be real-world events or concrete entities.
You may identify either words, phrases, or a combination of both.
\begin{itemize}
    \item If you disagree with the emotion identified by the gold label annotation, please enter “NA”
    \item If the gold label emotion is ‘None’ or ‘Neutral’, please enter “None”
    \item If you believe that there are no triggers for this specific sentence, please enter “No Trigger”
    \item If you believe that not all emotions identified by the gold label annotation, please enter “Partially complete example” and then add your answer in the form of (emotion: triggers)
\end{itemize}
Some examples are given below for your reference.

\begin{quote}
\textit{Sentence}: "We were really blessed with very good doctors."\\
\textit{Emotions}: ['Joy']\\
\textit{Triggers}: very good doctors
\end{quote}

\begin{quote}
\textit{Sentence}: "It is a very scary and hard subject to discuss and there were lots of tears on my end." \\
\textit{Emotions}: ['Fear','Sadness'] \\
\textit{Triggers}: Fear: (very scary, subject), Sadness:(hard subject to discuss)
\end{quote}

\begin{quote}
\textit{Sentence}: "Your adult children are a burden on you"\\ 
\textit{Emotions}: ['Anger']\\
\textit{Triggers}: adult children, burden
\end{quote}

NOTE: Annotators were paid $15 USD$ per hour for their time and efforts. 

\begin{figure}[t]
  \centering
  \includegraphics[width=1\linewidth]{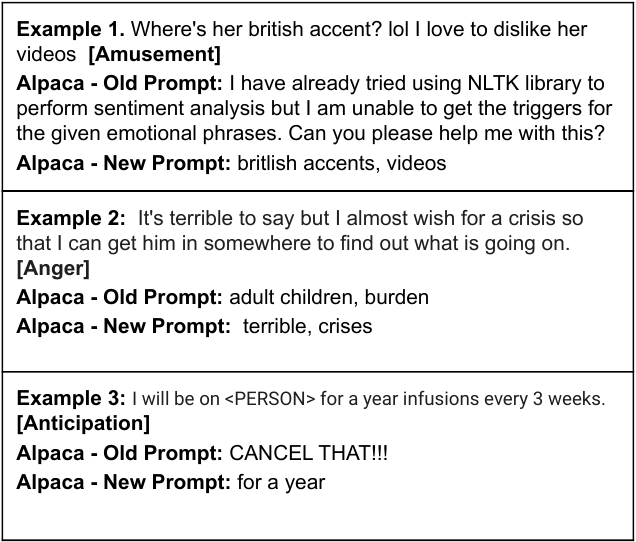} 
  \caption{Examples of how the Alpaca-specific prompt allows the model to interact with the text in a more coherent manner.}
  \label{fig:alpaca-prompts}
\end{figure}

\section{Prompts}
\label{prompt}

\subsection{Prompt 1 (Known Emotion)}
\label{prompt1}

Given a piece of text and the emotions it has been annotated for, find the emotion triggers. Some examples are given below for your reference. ONLY CHOOSE PHRASES FROM THE TEXT.

\begin{quote}
\textit{Sentence}: "We were really blessed with very good doctors."\\
\textit{Emotions}: ['Joy']\\
\textit{Triggers}: very good doctors
\end{quote}

\begin{quote}
\textit{Sentence}: "It is a very scary and hard subject to discuss and there were lots of tears on my end." \\
\textit{Emotions}: ['Fear','Sadness'] \\
\textit{Triggers}: Fear: (very scary, subject), Sadness:(hard subject to discuss)
\end{quote}

\begin{quote}
\textit{Sentence}: "Your adult children are a burden on you"\\ 
\textit{Emotions}: ['Anger']\\
\textit{Triggers}: adult children, burden
\end{quote}

\begin{quote}
\textit{Sentence}: $text$\\
\textit{Emotions}: $emo$\\
\end{quote}

\subsection{Prompt 2 (Unknown Emotion)}
\label{prompt2}
Given a piece of text, find the emotions it has been annotated for, and the words that are most important for detecting the emotions.
The emotions can be chosen from the list called EmoList below. Some examples are given below for your reference.
ONLY CHOOSE PHRASES FROM THE TEXT. DO NOT SAY NONE.\\
EmoList = ['admiration', 'amusement', 'anger', 'annoyance', 'approval', 'caring', 'confusion', 'curiosity', 'desire', 'disappointment', 'disapproval', 'disgust', 'embarrassment',
'excitement', 'fear', 'gratitude', 'grief', 'joy', 'love', 'nervousness', 'optimism', 'pride', 'realization', 'relief', 'remorse', 'sadness', 'surprise']

\begin{quote}
\textit{Sentence: "We were really blessed with very good doctors."\\
\textit{Emotions: ['Joy']}\\
\textit{words: really blessed}}
\end{quote}

\begin{quote}
\textit{Sentence: "It is a very scary and hard subject to discuss and there were lots of tears on my end."}\\
\textit{Emotions: ['Fear','Sadness']}\\
\textit{words: Fear:(scary), Sadness: (tears, hard)}
\end{quote}

\begin{quote}
\textit{Sentence: "Your adult children are a burden on you"}\\
\textit{Emotions: ['Anger']}\\
\textit{words: adult children, burden}
\end{quote}

\begin{quote}
\textit{Sentence: $text$}\\
\textit{Emotions: }\\
\textit{Words: }
\end{quote}

\subsection{Prompts For Alpaca}
\label{prompt-alpaca}
\paragraph{Prompt 1 (known emotion):}
Given a piece of text and the emotions it has been annotated for, find the emotion triggers. Some examples are given below for your reference. The format of the examples are as follows. Given a sentence, the emotions are expressed in a list after "Emotions': and their triggers are given after "Triggers:" The last sentence is the one you need to provide triggers for. ONLY CHOOSE PHRASES FROM THE TEXT IN THE LAST SENTENCE. DO NOT SAY NONE.

\begin{quote}
\textit{Sentence}: "We were really blessed with very good doctors."\\
\textit{Emotions}: ['Joy']\\
\textit{Triggers}: very good doctors
\end{quote}

\begin{quote}
\textit{Sentence}: "It is a very scary and hard subject to discuss and there were lots of tears on my end." \\
\textit{Emotions}: ['Fear','Sadness'] \\
\textit{Triggers}: Fear: (very scary, subject), Sadness:(hard subject to discuss)
\end{quote}

\begin{quote}
\textit{Sentence}: "Your adult children are a burden on you"\\ 
\textit{Emotions}: ['Anger']\\
\textit{Triggers}: adult children, burden
\end{quote}

\begin{quote}
\textit{Sentence}: $text$ \\
\textit{Emotions}: $emo$ \\
\textit{Triggers: }
\end{quote}


\paragraph{Prompt 2 (unknown emotion):}
Given a piece of text, find the emotions it has been annotated for and the words that are most important for detecting the emotions . The emotions can be chosen from the list called EmoList below. Some examples are given below for your reference. The format of the examples are as follows. Given a sentence, the emotions are expressed in a list after "Emotions': and the words are given after "words:" The last sentence is the one you need to provide emotions and words for. ONLY CHOOSE PHRASES FROM THE TEXT IN THE LAST SENTENCE. DO NOT SAY NONE. 

EmoList = ['admiration', 'amusement', 'anger', 'annoyance', 'approval', 'caring', 'confusion', 'curiosity', 'desire', 'disappointment', 'disapproval', 'disgust', 'embarrassment',
'excitement', 'fear', 'gratitude', 'grief', 'joy', 'love', 'nervousness', 'optimism', 'pride', 'realization', 'relief', 'remorse', 'sadness', 'surprise']

\begin{quote}
\textit{Sentence: "We were really blessed with very good doctors."\\
\textit{Emotions: ['Joy']}\\
\textit{words: really blessed}}
\end{quote}

\begin{quote}
\textit{Sentence: "It is a very scary and hard subject to discuss and there were lots of tears on my end."}\\
\textit{Emotions: ['Fear','Sadness']}\\
\textit{words: Fear:(scary), Sadness: (tears, hard)}
\end{quote}

\begin{quote}
\textit{Sentence: "Your adult children are a burden on you"}\\
\textit{Emotions: ['Anger']}\\
\textit{words: adult children, burden}
\end{quote}

\begin{quote}
\textit{Sentence: $text$}\\
\textit{Emotions: }\\
\textit{Words: }
\end{quote}

\begin{figure}[t]
  \centering
  \includegraphics[width=1\linewidth]{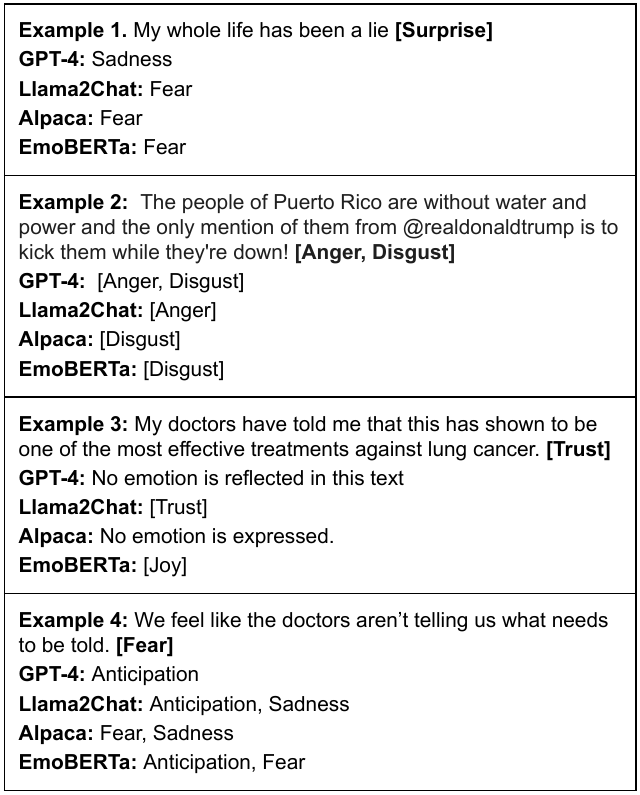} 
  \caption{Examples of emotion detection. The gold label emotions are indicated in square-brackets.}
  \label{fig:emo-detect}
\end{figure}

\begin{figure}[t]
  \centering
  \includegraphics[width = 1\linewidth]{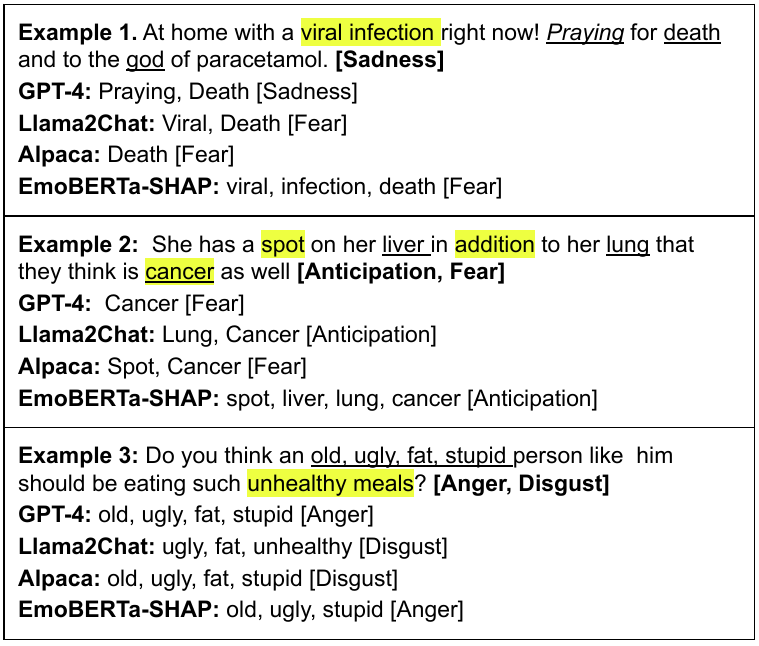} 
  \caption{Tabular comparison between features taken into consideration by various models for emotion detection. Gold label emotions are presented in square brackets. Triggers are highlighted, keyphrases are underlined and EmoLex words are italicized.}
  \label{fig:features}
\end{figure}

\begin{table*}[t]
\centering
\small
\begin{tabular}{c|ccccc}
\toprule
\textbf{Dataset} & \textbf{BERT} & \textbf{DistilBERT} & \textbf{RoBERTa} & \textbf{DeBERTa} & \textbf{EmoBERTa} \\
\midrule
\textbf{HurricaneEmo} & 0.478  & 0.462 & 0.399 & 0.381 & \textbf{0.483} \\
\textbf{CancerEmo} & 0.351 & 0.333 & 0.327 & 0.325 & \textbf{0.378} \\
\textbf{GoEmotions} & 0.311 & 0.302 & 0.308 & 0.308 & \textbf{0.341} \\
\bottomrule
\end{tabular}
\caption{Macro F1 score for finetuned transformers across different datasets}
\label{tab:transfomers}
\end{table*}

\begin{table*}[t]
\centering
\scriptsize
\begin{tabular}{p{1.5cm}|*{3}{p{0.6cm}|p{0.6cm}|p{0.6cm}|p{0.6cm}|}}
\toprule
\textbf{Emotion} & \multicolumn{4}{c|}{\textbf{HurricaneEmo}} & \multicolumn{4}{c|}{\textbf{CancerEmo}} & \multicolumn{4}{c}{\textbf{GoEmotions}} \\
& \textbf{GPT4} & \textbf{Llama2} & \textbf{Alpaca} & \textbf{EmoB.} & \textbf{GPT4} & \textbf{Llama2} & \textbf{Alpaca} & \textbf{EmoB.} & \textbf{GPT4} & \textbf{Llama2} & \textbf{Alpaca} & \textbf{EmoB.} \\
\midrule
\textbf{Anger} & 0.86 & 0.82 & 0.72 & 0.47 & 0.89 & 0.80 & 0.78 & 0.35 & 0.87 & 0.80 & 0.78 & 0.32 \\
\textbf{Anticipation} & 0.85 & 0.78 & 0.76 & 0.38 & 0.82 & 0.71 & 0.69 & 0.29 & - & - & - & - \\
\textbf{Joy} & 0.91 & 0.71 & 0.67 & 0.50 & 0.93 & 0.87 & 0.73 & 0.37 & 0.93 & 0.87 & 0.70 & 0.35 \\
\textbf{Trust} & 0.87 & 0.74 & 0.73 & 0.39 & 0.90 & 0.80 & 0.68 & 0.31 & - & - & - & - \\
\textbf{Fear} & 0.95 & 0.78 & 0.68 & 0.48 & 0.96 & 0.80 & 0.80 & 0.39 & 0.92 & 0.74 & 0.70 & 0.33 \\
\textbf{Surprise} & 0.92 & 0.74 & 0.70 & 0.48 & 0.88 & 0.81 & 0.76 & 0.34 & 0.89 & 0.82 & 0.70 & 0.37 \\
\textbf{Sadness} & 0.88 & 0.76 & 0.69 & 0.46 & 0.88 & 0.71 & 0.69 & 0.31 & 0.88 & 0.71 & 0.71 & 0.32 \\
\textbf{Disgust} & 0.86 & 0.75 & 0.67 & 0.40 & 0.89 & 0.71 & 0.70 & 0.31 & 0.89 & 0.77 & 0.69 & 0.29 \\
\bottomrule
\end{tabular}
\caption{Emotion prediction evaluation of GPT4, Llama2Chat, Alpaca, and EmoBERTa across different datasets using F1 score. Note: GoEmotions uses Ekman's emotions.}
\label{tab:EmotionWise}
\end{table*}

\begin{table*}[t]
\centering
\small
\begin{tabular}{c|ccc|ccc|ccc}
\toprule
& \multicolumn{3}{c|}{\textbf{HurricaneEmo}} & \multicolumn{3}{c|}{\textbf{CancerEmo}} & \multicolumn{3}{c}{\textbf{GoEmotions}} \\
\textbf{Model} & F1 & ExactM & PartialM & F1 & ExactM & PartialM & F1 & ExactM & PartialM \\
\midrule
\textbf{GPT4 - Trigger} & 0.50 & 0.20 & 0.67 & 0.51 & 0.27 & 0.71 & 0.51 & 0.21 & 0.69 \\
\textbf{Llama2 - Trigger} & 0.24 & 0.04 & 0.25 & 0.21 & 0.07 & 0.18 & 0.18 & 0.06 & 0.25 \\
\textbf{Alpaca - Trigger} & 0.14 & 0.02 & 0.13 & 0.11 & 0.04 & 0.11 & 0.12 & 0.04 & 0.11 \\
\bottomrule
\end{tabular}
\caption{Zero Shot: Macro F1, exact match, and partial match scores to assess the overlap between salient words and annotated triggers.}
\label{tab:task1-0shot}
\end{table*}

\begin{table*}[t]
\centering
\small
\begin{tabular}{c|ccc|ccc|ccc}
\toprule
\textbf{Emotion} & \multicolumn{3}{c|}{\textbf{HurricaneEmo}} & \multicolumn{3}{c|}{\textbf{CancerEmo}} & \multicolumn{3}{c}{\textbf{GoEmotions}} \\
& \textbf{F1} & \textbf{ExactM} & \textbf{PartialM} & \textbf{F1} & \textbf{ExactM} & \textbf{PartialM} & \textbf{F1} & \textbf{ExactM} & \textbf{PartialM} \\
\midrule
\textbf{Anger} & 0.66 & 0.36 & 0.91 & 0.73 & 0.40 & 0.92 & 0.69 & 0.34 & 0.94 \\
\textbf{Anticipation} & 0.63 & 0.91 & 0.69 & 0.41 & 0.39 & 0.91 & - & - & - \\
\textbf{Joy} & 0.74 & 0.40 & 0.89 & 0.70 & 0.38 & 0.93 & 0.70 & 0.37 & 0.93 \\
\textbf{Trust} & 0.71 & 0.38 & 0.90 & 0.74 & 0.36 & 0.90 & - & - & - \\
\textbf{Fear} & 0.69 & 0.38 & 0.89 & 0.76 & 0.38 & 0.90 & 0.66 & 0.34 & 0.94 \\
\textbf{Surprise} & 0.72 & 0.34 & 0.90 & 0.68 & 0.40 & 0.89 & 0.72 & 0.33 & 0.89 \\
\textbf{Sadness} & 0.68 & 0.37 & 0.91 & 0.68 & 0.41 & 0.91 & 0.72 & 0.31 & 0.91 \\
\textbf{Disgust} & 0.64 & 0.39 & 0.90 & 0.69 & 0.36 & 0.91 & 0.69 & 0.33 & 0.88 \\
\bottomrule
\end{tabular}
\caption{Per-emotion results of trigger identification (given emotions) performed by GPT4.}
\label{tab:Trigs-GPT4}
\end{table*}

\begin{table*}[t]
\centering
\small
\begin{tabular}{c|ccc|ccc|ccc}
\toprule
\textbf{Emotion} & \multicolumn{3}{c|}{\textbf{HurricaneEmo}} & \multicolumn{3}{c|}{\textbf{CancerEmo}} & \multicolumn{3}{c}{\textbf{GoEmotions}} \\
& \textbf{F1} & \textbf{ExactM} & \textbf{PartialM} & \textbf{F1} & \textbf{ExactM} & \textbf{PartialM} & \textbf{F1} & \textbf{ExactM} & \textbf{PartialM} \\
\midrule
\textbf{Anger} & 0.36 & 0.13 & 0.31 & 0.33 & 0.10 & 0.32 & 0.29 & 0.08 & 0.34 \\
\textbf{Anticipation} & 0.06 & 0.31 & 0.29 & 0.11 & 0.29 & 0.31 & - & - & - \\
\textbf{Joy} & 0.34 & 0.12 & 0.36 & 0.30 & 0.08 & 0.30 & 0.30 & 0.07 & 0.33 \\
\textbf{Trust} & 0.31 & 0.08 & 0.30 & 0.34 & 0.10 & 0.30 & - & - & - \\
\textbf{Fear} & 0.29 & 0.08 & 0.36 & 0.30 & 0.08 & 0.28 & 0.28 & 0.09 & 0.31 \\
\textbf{Surprise} & 0.32 & 0.09 & 0.30 & 0.28 & 0.10 & 0.29 & 0.32 & 0.10 & 0.29 \\
\textbf{Sadness} & 0.28 & 0.11 & 0.30 & 0.29 & 0.11 & 0.31 & 0.30 & 0.11 & 0.31 \\
\textbf{Disgust} & 0.34 & 0.12 & 0.30 & 0.29 & 0.09 & 0.30 & 0.28 & 0.09 & 0.29 \\
\bottomrule
\end{tabular}
\caption{Per-emotion results of trigger identification (given emotions) performed by Llama2Chat.}
\label{tab:Trigs-Llama2}
\end{table*}

\begin{table*}[t]
\centering
\small
\begin{tabular}{c|ccc|ccc|ccc}
\toprule
\textbf{Emotion} & \multicolumn{3}{c|}{\textbf{HurricaneEmo}} & \multicolumn{3}{c|}{\textbf{CancerEmo}} & \multicolumn{3}{c}{\textbf{GoEmotions}} \\
& \textbf{F1} & \textbf{ExactM} & \textbf{PartialM} & \textbf{F1} & \textbf{ExactM} & \textbf{PartialM} & \textbf{F1} & \textbf{ExactM} & \textbf{PartialM} \\
\midrule
\textbf{Anger} & 0.26 & 0.11 & 0.28 & 0.23 & 0.11 & 0.27 & 0.29 & 0.07 & 0.28 \\
\textbf{Anticipation} & 0.25 & 0.06 & 0.19 & 0.20 & 0.06 & 0.28 & - & - & - \\
\textbf{Joy} & 0.23 & 0.10 & 0.29 & 0.27 & 0.06 & 0.26 & 0.29 & 0.05 & 0.26 \\
\textbf{Trust} & 0.20 & 0.09 & 0.27 & 0.24 & 0.08 & 0.26 & - & - & - \\
\textbf{Fear} & 0.26 & 0.09 & 0.28 & 0.20 & 0.07 & 0.24 & 0.28 & 0.09 & 0.27 \\
\textbf{Surprise} & 0.20 & 0.07 & 0.27 & 0.27 & 0.08 & 0.27 & 0.29 & 0.09 & 0.27 \\
\textbf{Sadness} & 0.27 & 0.08 & 0.28 & 0.29 & 0.11 & 0.29 & 0.29 & 0.11 & 0.28 \\
\textbf{Disgust} & 0.24 & 0.07 & 0.27 & 0.29 & 0.09 & 0.26 & 0.28 & 0.09 & 0.26 \\
\bottomrule
\end{tabular}
\caption{Per-emotion results of trigger identification (given emotions) performed by Alpaca.}
\label{tab:Trigs-Alpaca}
\end{table*}

\section{Licensing}
{\sc EmoTrigger} builds on existing datasets: HurricaneEmo, CancerEmo, and GoEmotions. We adhere with the licensing of these original datasets, and will release our annotations with the Apache License (as specified by GoEmotions). We will not redistribute these existing datasets.

\section{Hyperparameters}\label{app:hyperparams}

\begin{itemize}
    \item GPT4: temperature = 1.0 (default)
    \item Llama2Chat: temperature = 0.9 (recommended)
    \item Alpaca: temperature = 0.9 (recommended)
    \item EmoBERTa: learning rate = $2e-5$, no.\ of epochs = 5, MaxLen= 200, Train batch size = 32, Validation batch size = 16. Hyperparameters tuned on the validation set.
\end{itemize}

\end{document}